\documentclass[sigconf]{acmart}
\AtBeginDocument{%
  }

\setcopyright{acmlicensed}
\copyrightyear{2025}
\acmYear{2025}
\acmDOI{XXXXXXX.XXXXXXX}
\acmConference[AIR 2025]{Advances In Robotics - 7th International Conference of The Robotics Society
}{July 02-05,
  2025} {IIT Jodhpur, India}
\acmISBN{978-1-4503-XXXX-X/2018/06}

\newenvironment{rcases}
{\left.\begin{aligned}}
	{\end{aligned}\right\rbrace}
\usepackage{subcaption}
\usepackage{bm}
\copyrightyear{2025}
\acmYear{2025}
\setcopyright{cc}
\setcctype{by}
\acmConference[AIR 2025]{Advances in Robotics 2025}{July 02--05, 2025}{Jodhpur, India}
\acmBooktitle{Advances in Robotics 2025 (AIR 2025), July 02--05, 2025, Jodhpur, India}
\acmPrice{}
\acmDOI{10.1145/3787370.3787419}
\acmISBN{979-8-4007-1383-5/2025/07}
\begin{document}

\title{TO-SoFiT: Topology Optimization of Hydraulic Soft Fish Tail Design for programmable undulating locomotion}

\author{A Padmaprabhan}
\affiliation{%
  \institution{Indian Institute of Technology Hyderabad}
  \state{Telangana, 502285}
  \country{India}
}

\author{Amal Shaji }
\affiliation{%
  \institution{Indian Institute of Technology Hyderabad}
  \state{Telangana, 502285}
  \country{India}
}

\author{Prabhat Kumar}
\orcid{1234-5678-9012}
\authornotemark[1]
\affiliation{%
  \institution{Indian Institute of Technology Hyderabad}
  \state{Telangana, 502285}
  \country{India}
  }
  \email{pkumar@mae.iith.ac.in}

\begin{abstract}
Soft robots leverage compliant materials to generate motion through controlled elastic deformation, making them ideal for delicate tasks such as underwater exploration and biomimetic marine systems. Although hydraulic/pneumatic actuation remains pivotal for such systems, the lack of systematic design frameworks has hindered the development of robots capable of complex 3D motion, such as fish-like swimming. This work introduces a topology optimization method to automate the design of a hydraulic soft fish tail, explicitly addressing the design-dependent coupling between fluidic actuation and structural deformation. We use a Darcy law-based model augmented with a drainage term to simulate spatially varying hydraulic pressure loads, translating these into consistent nodal forces via finite element analysis. The employed robust multi-criteria optimization formulation balances deformation efficiency, fluid-structure interaction, geometric manufacturability, and required stiffness for optimizing a bioinspired soft fish tail for 3D swimming kinematics. The optimized tail topology is incorporated into a pneumatic network actuator and computationally validated under various hydraulic loads, achieving tunable undulatory amplitudes and multiaxis bending for depth adjustment. The optimized 2D tail outperforms its rectangular counterpart. By cascading optimized tail segments, we demonstrate programmable swimming patterns in soft robotic fish tails at different hydraulic loads. This work advances the systematic codesign of hydraulic actuators and soft structures, offering a pathway to automate underwater robots with optimized design and vertebrate-like agility in confined aquatic environments. Our implementations and simulations are publicly available at \href{https://github.com/PrabhatIn/TO-SoFiT}{\textbf{\texttt{TO-SoFiT}}} repository.
\end{abstract}

\begin{CCSXML}
<ccs2012>
<concept>
<concept_id>10010147.10010341.10010370</concept_id>
<concept_desc>Computing methodologies~Simulation evaluation</concept_desc>
<concept_significance>500</concept_significance>
</concept>
<concept>
<concept_id>10010147.10010341.10010342.10010343</concept_id>
<concept_desc>Computing methodologies~Modeling methodologies</concept_desc>
<concept_significance>500</concept_significance>
</concept>
<concept>
<concept_id>10010405.10010481.10010483</concept_id>
<concept_desc>Applied computing~Computer-aided manufacturing</concept_desc>
<concept_significance>500</concept_significance>
</concept>
<concept>
<concept_id>10010520.10010553.10010554.10010557</concept_id>
<concept_desc>Computer systems organization~Robotic autonomy</concept_desc>
<concept_significance>300</concept_significance>
</concept>
</ccs2012>
\end{CCSXML}

\ccsdesc[500]{Computing methodologies~Simulation evaluation}
\ccsdesc[500]{Computing methodologies~Modeling methodologies}
\ccsdesc[500]{Applied computing~Computer-aided manufacturing}
\ccsdesc[300]{Computer systems organization~Robotic autonomy}

\keywords{Soft robotic fish, Topology optimization, Design-dependent loads, Hydraulic actuation, Programmable 3D swimming}

\maketitle

\section{Introduction}
Soft robots are constructed from flexible materials and feature lightweight monolithic designs~\cite{xavier2022soft,kumar2022towards}. They are actuated using pneumatic or hydraulic (fluidic pressure) loads and rely on motion generated by elastic deformation to perform their tasks~\cite{kumar2022towards}. Soft robotic fish have numerous underwater applications, such as enabling non-invasive monitoring of fragile reef ecosystems. The soft fish (SoFi)  platform~\cite{katzschmann2018exploration} captures high-resolution images at depths of up to 18 meters using a fisheye lens. These soft robotics fish transcend traditional robotics by integrating biological principles of undulatory fish with engineered systems. As material innovations and autonomy algorithms advance, these platforms will be necessary for sustainable underwater exploration, assessment, and testing, offering solutions where conventional techniques fail~\cite{VANDENBERG2022e00320}. Therefore, interest in designing them for different underwater applications is constantly growing. However, these robots are generally created manually using heuristic methods that greatly depend on designers' experience, skills, and knowledge. Therefore, this paper aims to present a systematic approach using topology optimization to design a pneumatically actuated soft robotic fish tail, a primary propulsive component of many underwater robots.

Topology optimization (TO), a numerical technique, provides an optimized material distribution within a given design domain to achieve a specific objective under physical/geometrical constraints. A typical TO approach discretizes the design domain using finite elements (FEs) to solve the relevant boundary value problem~\cite{kumar2023honeytop90}. For the optimization process, each element is assigned a design variable $\rho \in[0,\,1]$, and the optimization problem is relaxed ($0<\rho<1$)~\cite{kumar2023honeytop90}. $\rho = 1$ and $\rho = 0$ indicate element's solid and void states, respectively. The hydraulic pressure load is a design-dependent force that changes its location, magnitude, and/or direction as TO progresses~\cite{kumar2020topology,kumar2020topology3Dpressure,kumar2023topress}. 

The behavior of a soft robot closely resembles that of a compliant mechanism (CM), where motion is achieved through the deformation of flexible components under constant or design-dependent pressure actuations~\cite{kumar2022towards,kumar2022topological}. Modeling the latter type of actuation introduces several distinct challenges within a TO framework~\cite{kumar2020topology,kumar2023topress}, which become even more pronounced when designing CMs with such loads~\cite{kumar2020topology,kumar2022topological}. Some approaches to creating such pressure-actuated CMs exist and can be found in~\cite{kumar2020topology,kumar2020topology3Dpressure,vasista2012design,de2020topology,lu2021topology,kumar2022topological,kumar2024sorotop}. In this work, we adopt the method proposed in~\cite{kumar2020topology,kumar2020topology3Dpressure,kumar2023topress}, which utilizes Darcy's law with a conceptualized drainage term to model pressure loads. This innovative approach, built on the standard finite element method, offers an efficient way to relate pressure loads to design variables, implicitly defines the pressure loading surface, and enables the determination of load sensitivities using the adjoint variable method~\cite{kumar2020topology,kumar2020topology3Dpressure,kumar2023topress,kumar2022topological,kumar2024sorotop}.

Developing hydraulically actuated soft fish tails for undulating locomotion presents unique challenges similar to compliant mechanism design, requiring precise coordination between fluid-driven deformation and pressure-loaded chamber design. Although researchers have now started using TO for designing soft robotics~\cite {kumar2022towards,pinskier2023automated,pinskier2024diversity}, some previous approaches~\cite{hiller2011automatic,chen2018topology,zhang2018design} neglected the critical \textit{design-dependent} hydraulic interactions essential for underwater propulsion, which alter optimized designs. Assuming fixed-pressure load orientations can fail to account for the spatial hydraulic coupling inherent in soft-robotic fish-tail kinematics, where fluidic pressures are redistributed across deforming PneuNets~\cite{kumar2022towards}. PneuNet or fluidic elastomeric actuators consist of chambers embedded in elastomeric structures and have been important in soft robotics~\cite{shepherd2011multigait,Mosadegh2014,kumar2022towards}. Generally, the pneumatic network consists of an isotropic elastomer material with a series of chambers that extend, bend, or twist into different shapes when pressurized with a fluid pressure load.

We optimize the soft fish tail as a bioinspired CM, following Katzschmann et al.~\cite{katzschmann2018exploration}, which converts hydraulic pressure into controlled bending motions via a PneuNet of pressure chambers. The design-dependent nature of the hydraulic load is modeled using the Darcy law~\cite{kumar2022topological}. A symmetric half of the soft fish tail's PneuNet is optimized as a CM design to maximize the output displacement. \texttt{SoRoTop} based code~\cite{kumar2024sorotop} that uses the Darcy method to design centrally pressurized structures, is used for optimizing the symmetric half of one unit of soft fish tail.

The remainder of the paper is structured as follows. Sec.~\ref{Sec:Sot_tail_undulating} details about the soft tail design for locomotion. Sec.~\ref{Sec:TOPopt_formulation} provides pressure load modeling using the Darcy law in brief and the TO formulation. Sec.~\ref{Sec:Fishtaildesign} presents an optimized design for a PneuNet soft robotic fish tail member. The optimized design is extracted, and its cross-sections are used for analysis. The CAD model consists of a design with serial chambers consisting of optimized structures. Nonlinear finite element analyses are performed in \texttt{ABAQUS}. Lastly, the conclusions are drawn in Sec.~\ref{Sec:Con}.

\section{Soft tail design for undulating locomotion}\label{Sec:Sot_tail_undulating}

The tail design mimics a caudal fin and the posterior peduncle of a fish, consisting of a rib structure that expands and contracts by the thin exterior skin under hydraulic actuation. The evenly spaced chambers, which resemble a rib-like structure of the array of PneuNets, are connected by a central channel and can be pressurized or depressurized. This rib-like structure is present throughout the entire caudal fin of the soft robotic fish, allowing it to bend repeatedly under fluidic pressure. After each actuation cycle, its inherent elasticity forces it back to its original state.
The undulating movement is achieved by alternating fluidic flow into the two cavities. 
Actuation is achieved by an external gear pump operating at a desired frequency, whose outlets are connected to the two pneumatic chambers to allow fluid transport between them. The flexing actuation of the tail is achieved by alternating the flow with an external gear pump actuator at a desired frequency. Next, we describe the TO framework for generating the optimized fish tail.

\section{Topology optimization framework}\label{Sec:TOPopt_formulation}
We employ a density-based TO, wherein the modified SIMP method is used to interpolate Young's modulus as
\begin{equation}\label{Eq:SIMP}
	\mathrm{E}_i = \mathrm{E}_\mathrm{min} + \bar{\rho}^p_i(\mathrm{E}_1-\mathrm{E}_\mathrm{min}),
\end{equation}
where $E_i$ indicates the interpolated Young's modulus of element $i$. Interpolation is performed using Young's modulus of the void state, $E_0$, and that of the solid state, $E_1$. $p=3$ is the penalty parameter. $\bar{\rho}_i$, physical variable, is the projected design variable obtained as
\begin{equation}\label{Eq:projectionFilt}
	\bar{\rho}_i = \mathcal{H}(\tilde{{\rho_i}},\,\beta,\,\eta)=\frac{\tanh(\beta \eta) + \tanh(\beta(\tilde{\rho}_i-\eta))} {\tanh{\left(\beta \eta\right)}+\tanh{\left(\beta(1 - \eta)\right)}}, 
\end{equation}
where $\left\{\eta,\,\beta\right\}$ are projection parameters that define the transition point and steepness, respectively. The latter is varied in a continuation fashion to obtain a solution close to 0-1. $\tilde{{\rho_i}}$, the filtered variable, is determined as
\begin{equation} \label{Eq:density_filter}
	\tilde{\rho}_i = \frac{\sum_{j=1}^{{nel}}\rho_j {v}_j {w}(\mathbf{x}_i,\mathbf{x}_j)}{\sum_{j=1}^{{nel}} {v}_j {w}(\mathbf{x}_i,\mathbf{x}_j) } \; ,
\end{equation}
where weight ${w}(\mathbf{x}_i,\mathbf{x}_j)$ = $\mathrm{max}  \left(0 \; , \; 1-\frac{\| \mathrm{\mathbf{x}}_i - \mathrm{\mathbf{x}}_j \|}{\mathrm{r}_\mathrm{fill}} \right)$ \cite{bruns2001topology}, with the filter radius $\mathrm{r}_\mathrm{fill}$. ${v}_j$ and $nel$ indicate the volume of element~$j$ and the total number of elements used to parameterize the domain. 
\begin{figure}[h!]
	\centering
	\includegraphics[scale = 0.95]{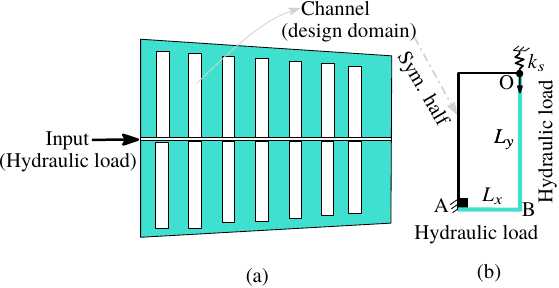}
	\caption{(a) A schematic diagram of fish soft tail with channels~\cite{katzschmann2018exploration}. (b) A symmetric half-domain design  is used to optimize channel design. $k_s$ indicates stiffness for the output spring. Hydraulic pressure loads are applied on AB and BO. BO is the symmetric edge, and point O is the output location. $\frac{L_x}{L_y} = 1:4$.} \label{fig:section_domain}
\end{figure}

\begin{figure}[h!] 
	\begin{subfigure}{0.21\textwidth}
		\centering
		\includegraphics[scale = 0.4]{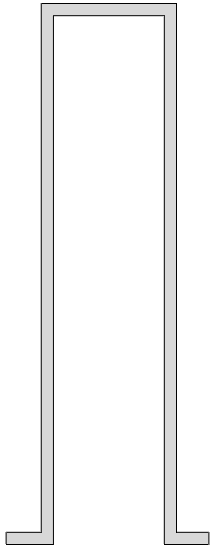}
		\caption{Rectangular fish tail array unit}
	\end{subfigure}
	\quad
	\begin{subfigure}{0.21\textwidth}
		\centering
		\includegraphics[scale =0.40]{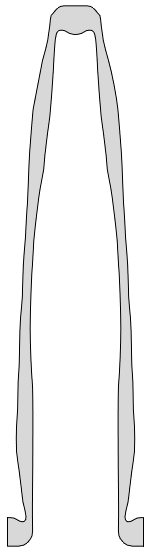}
		\caption{Optimized fish tail array unit}
	\end{subfigure}
	\caption{Cross-sectional views of trivial and optimized PneuNet chambers}
	\label{fig:results_2D CAD Model}
\end{figure}

\begin{figure*}[h!] 
	\begin{subfigure}{0.45\textwidth}
		\centering
		\includegraphics[scale = 0.6]{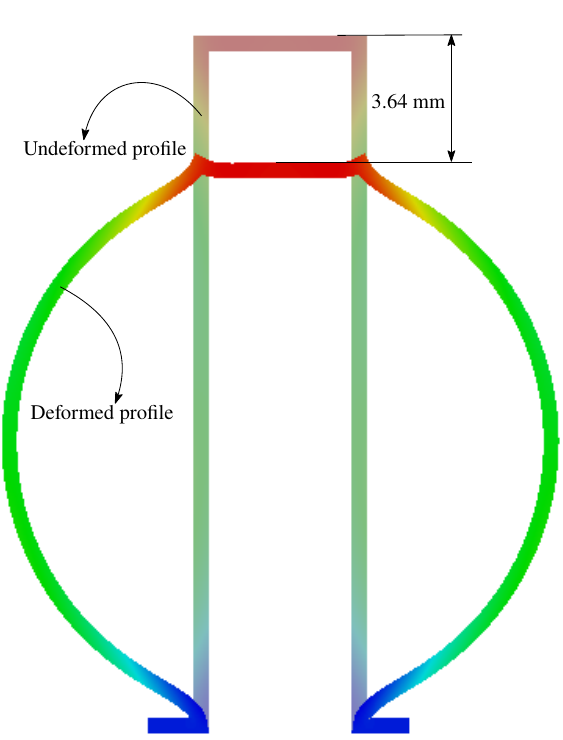}
		\caption{Rectangular chamber deformation}
	\end{subfigure}
	\quad
	\begin{subfigure}{0.45\textwidth}
		\centering
		\includegraphics[scale =0.6]{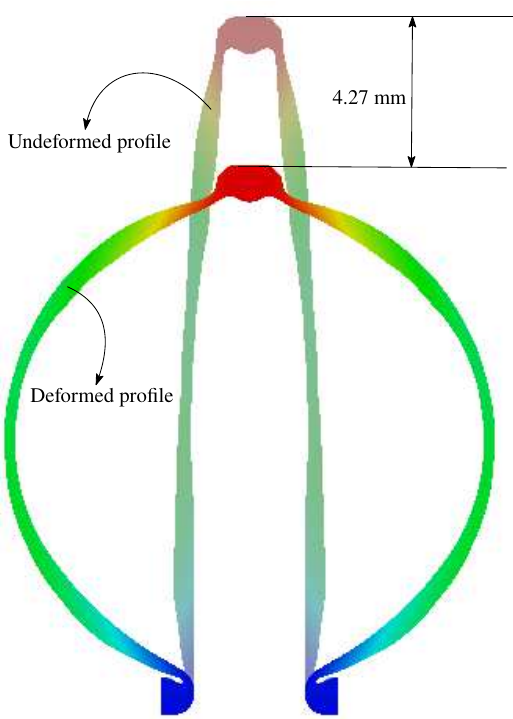}
		\caption{Optimized chamber deformation}
	\end{subfigure}
	\caption{Deformation profiles comparison between the 2D rectangular and the optimized fish tail units. Analysis is performed with \texttt{ABAQUS} with a 3kPa pressure load applied inside the chambers. The optimized tail deforms 17.30 \%  more than the rectangular counterpart, under similar pressure loading conditions. \textit{Color scheme}: Red and blue indicate maximum and minimum deformation, respectively.}
	\label{fig:results_2D results}
\end{figure*}

\subsection{Hydraulic pressure load modeling}
As mentioned above, we adopt the Darcy approach~\cite{kumar2020topology,kumar2020topology3Dpressure,kumar2023topress,kumar2022topological} to model hydraulic pressure actuation in the TO setting, which is briefly discussed for completeness. 
One determines the Darcy flux $\bm{q}$ in terms of pressure gradient $\nabla p$, permeability of the medium $\kappa$, and viscosity $\mu$ as~\cite{kumar2020topology}, as a function of every element in the design domain:
\begin{equation}\label{Eq:Darcyflux}
	\bm{q} = -\frac{\kappa}{\mu}\nabla p = -\mathcal{K}(\bar{\bm{\rho}}) \nabla p,
\end{equation}
where $K(\bar{\bm{\rho}})$ represents flow coefficient, which for element $i$ is defined as~~\cite{kumar2020topology}
\begin{equation}\label{Eq:Flowcoefficient}
	\mathcal{K}(\bar{\rho_i}) = \mathcal{K}_\text{v}\left(1-(1-\epsilon) \mathcal{H}(\bar{{\rho_i}},\,\beta_\kappa,\,\eta_\kappa)\right),
\end{equation}
where flow contrast $\epsilon = \frac{\mathcal{K}_\text{s}}{\mathcal{K}_\text{v}}$. $\mathcal{K}_\text{s}$ and $\mathcal{K}_\text{v}$ indicate the flow coefficient for the solid and void states of element~$i$, and \(\mathcal{H}(\bar{\rho})\) is the smooth Heaviside function.  $\left\{\eta_\kappa,\,\beta_\kappa\right\}$ parameters for the flow coefficient. Per~\cite{kumar2020topology,kumar2020topology3Dpressure,kumar2023topress} to ensure valid pressure variation in a TO setting, a drainage term $Q_\text{drain}$ must be added. $Q_\text{drain}= -D(\bar{\rho_i}) (p - p_{\text{ext}})$ with $D(\bar{\rho_i}) =  D_{\text{s}}\mathcal{H}(\bar{{\rho_e}},\,\beta_\text{d},\,\eta_\text{d})$. $\left\{\eta_\text{d},\,\beta_\text{d} \right\}$ are the parameters for $Q_\text{drain}$. With $Q_\text{drain}$,  the following balance equation is obtained~\cite{kumar2020topology}:
\begin{equation}\label{Eq:PDEstrong}
\nabla \cdot ( K \nabla p ) + Q_\text{drain} = 0 \quad \text{in } \Omega.
\end{equation}
Multiplying by a test function $w$ and integrating over $\Omega$ gives
\begin{equation}
\int_\Omega w \left( \nabla \cdot (K \nabla p) + D_s\mathcal{H}(\bar{\rho}) p \right) \, d\Omega = 0.
\end{equation}
Integrating the diffusion term by parts and assuming natural boundary conditions yields the weak form as
\begin{equation}\label{Eq:weakform}
\int_\Omega K \nabla w \cdot \nabla p \, d\Omega
+ \int_\Omega D_s \mathcal{H}(\bar{\rho}) w p \, d\Omega
= 0 \qquad \forall w.
\end{equation}
Approximating $p$ using finite element bi-linear shape functions $N_j|_{1,\,2,\,3,\,4}$ herein,
\[
p \approx \sum_j N_j p_j, \qquad w = N_i,
\]
leads to
\begin{equation}
\sum_j p_j \left[
\int_\Omega K \nabla N_i \cdot \nabla N_j \, d\Omega
+ \int_\Omega D_s \mathcal{H}(\bar{\rho}) N_i N_j \, d\Omega
\right] = 0 \quad \forall i.
\end{equation}
the system can be written as
\begin{equation}\label{Eq:PDEsolutionpressure}
(\mathbf{K}_p + \mathbf{K}_{Dp})\mathbf{p} = \mathbf{0}
\;\;\Longleftrightarrow\;\;
\mathbf{A}\mathbf{p} = \mathbf{0},
\end{equation}
while assuming the dynamic external pressure, $p_\text{ext}=0$ for surface propulsion, which can be modified according to the pressure at depths where the soft tail can be actuated, and surface flux equals zero. $\mathbf{A}$ and $\mathbf{p}$ are the global flow matrix and pressure vector, respectively. The obtained pressure field, $\mathbf{p}$, is converted into the global nodal force as~\cite{kumar2020topology}
\begin{equation}\label{Eq:nodalforce}
	\mathbf{F} = -\mathbf{T}\mathbf{p},
\end{equation}
where $\mathbf{T}$ is the global transformation matrix. The readers are referred to Ref.~\cite{kumar2020topology} for more details.

\subsection{Optimization formulation}
The robust formulation is used for optimization~\cite{wang2011projection}, in which a min-max objective is formulated. The permitted resource constraint is applied to the blueprint design, and the strain constraint to the eroded design~\cite{kumar2024sorotop,kumar2023topology}. The latter constraint helps to achieve a realizable/feasible optimized design to sustain the applied hydraulic load~\cite{kumar2024sorotop}. Mathematically, 
\begin{equation}\label{Eq:Optimizationequation}
	\begin{rcases}
		\underset{\bar{\bm{\rho}}(\tilde{\bm{\rho}}(\bm{\rho}))}{\text{min}:}
		f_0 = \max_{r}\, {u^\text{out}_r} = \max_{r}\left\{\bm{l}^\top \mathbf{u}_r\right\}|_{r =e,\,b}\\
		\text{such that:} \\
		\,{^1}{\bm{\lambda}_r}:\,\, \mathbf{A}_r\mathbf{p}_r = \mathbf{0 }\\
		\,{^2}\bm{\lambda}_r:\,\,  \mathbf{K}_r\mathbf{u}_r = \mathbf{F}_r = -\mathbf{T} \mathbf{p}_r\\
		\,{\Lambda}_1:\,\, \text{g}_1 \equiv \frac{V_b}{V^*}-1\le 0\\
		\,{\Lambda}_2:\,\,  \text{g}_2\equiv \frac{SE_e}{SE^*}-1 \le 0\\
		\, 0\le \rho_i,\,\tilde{\rho}_i,\,\bar{\rho}_i\le 1 \,\,\forall i
	\end{rcases},
\end{equation}
where quantities with subscript $e$ and $b$ indicate the eroded and blueprint designs, respectively. The global displacement vector and stiffness matrix are denoted as $\mathbf{u}_r$ and $\mathbf{K}_r$, respectively. Output displacement in the required direction is represented by  $u^\text{out}_r$. $\bm{l}$ consists of zeros except for the entries corresponding to the output degrees of freedom.
$V^*$ and $V_b$  denote the blueprint design's permitted and current volume fractions, respectively. Likewise, $SE_e$ and $SE^*$ represent the eroded design's current and prescribed strain energy. 

We use the gradient-based optimizer, \texttt{MMA} (Method of Moving Asymptotes)~\cite{svanberg1987method}, to update the design variables. The adjoint-variable method is used to determine the derivatives with respect to the physical variables. Therefore, the derivatives of the objective and constraints with respect to the design variables are determined using the chain rule as per~\cite{kumar2024sorotop}.

\section{Optimized Soft Fish Tail  Design}\label{Sec:Fishtaildesign}
We optimize the soft fish tail using the TO method presented above. 
The soft fish tail optimization design domain is inspired by its structure mentioned in~\cite{katzschmann2018exploration}. 
Only a symmetric part of one unit (Fig.~\ref{fig:section_domain}a) is optimized, as a CM that is depicted in Fig.~\ref{fig:section_domain}b.

The symmetric design domain (Fig.~\ref{fig:section_domain}b) is parameterized using $40\times 320$ bilinear finite elements. The volume fraction for the blueprint design and strain energy fraction for the eroded design are set to 0.2 and 0.80, respectively. The \texttt{SIMP} (Solid Isotropic Material with Penalization) penalty parameter $p=3$ is taken. $r_\text{fill} = 5.6$. The stiffness of the workpiece $k_s =1$ is set. $\left\{\eta_\kappa, \,\beta_\kappa\right\}$ = $\left\{\eta_d, \,\beta_d\right\}$ = $\left\{0.1,\,10\right\}$ are taken~\cite{kumar2024sorotop}. The load sensitivities are considered. The maximum value of the projection parameter $\beta$ is set to 128. The maximum number of iterations for the \texttt{MMA} optimizer is set to 400. The above inputs are given as inputs to the $\texttt{SoRoTop}$ solver to carry out the optimization of the symmetric half design domain (Fig.~\ref{fig:section_domain}).

\begin{figure}[h!]
	\centering
	\includegraphics[scale = 0.35]{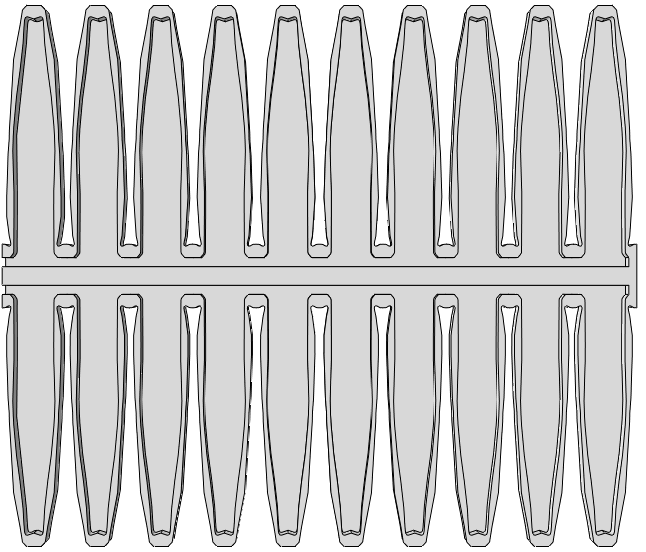}
	\caption{Cross-sectional view of the optimized soft elastomeric tail, containing parallel arrays of topology optimized PneuNets, separated by a stiffer strain-limiting layer.} \label{fig:full_3D_CAD_Model}
\end{figure}

\begin{figure*}
	\centering

	% ------------------ 5 kPa ------------------
	\begin{subfigure}{0.65\textwidth}
		\centering
		\begin{subfigure}{0.48\textwidth}
			\centering
			\includegraphics[width=\linewidth]{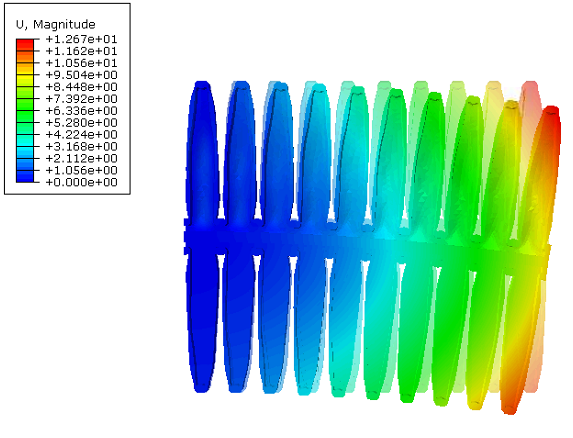}
			\caption*{Top chamber is activated}
		\end{subfigure}
		\hfill
		\begin{subfigure}{0.48\textwidth}
			\centering
			\includegraphics[width=\linewidth]{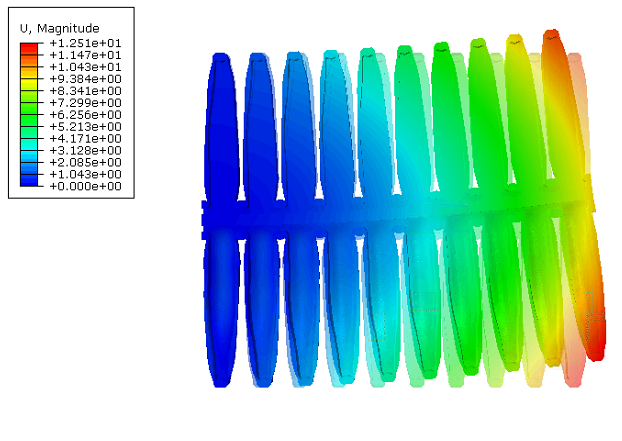}
			\caption*{Bottom chamber is activated}
		\end{subfigure}
		\caption{5 kPa}
	\end{subfigure}
	\hfill

	% ------------------ 10 kPa ------------------
	\begin{subfigure}{0.65\textwidth}
		\centering
		\begin{subfigure}{0.48\textwidth}
			\centering
			\includegraphics[width=\linewidth]{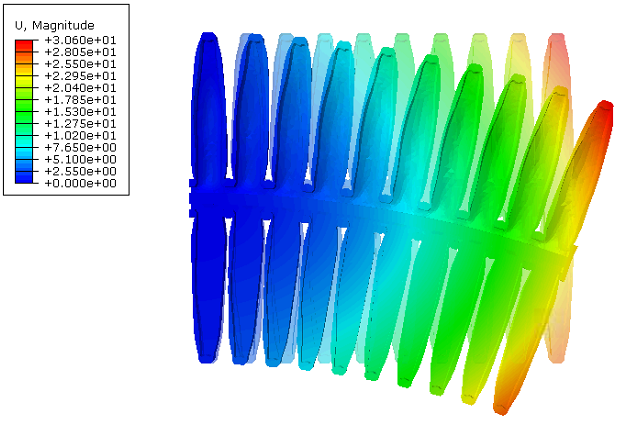}
			\caption*{Top chamber is activated}
		\end{subfigure}
		\hfill
		\begin{subfigure}{0.48\textwidth}
			\centering
			\includegraphics[width=\linewidth]{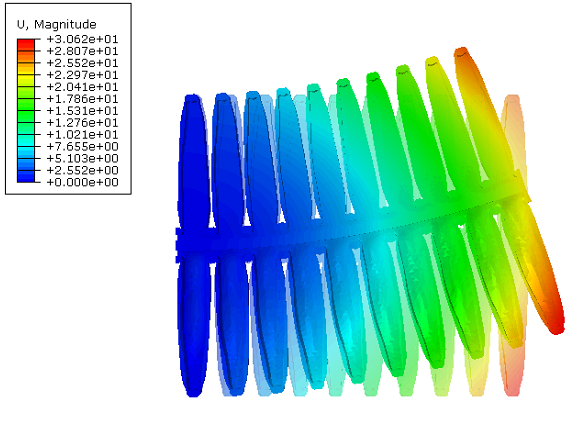}
			\caption*{Bottom chamber is activated}
		\end{subfigure}
		\caption{10 kPa}
	\end{subfigure}

	\vspace{0.5cm}

	% ------------------ 15 kPa ------------------
	\begin{subfigure}{0.65\textwidth}
		\centering
		\begin{subfigure}{0.48\textwidth}
			\centering
			\includegraphics[width=\linewidth]{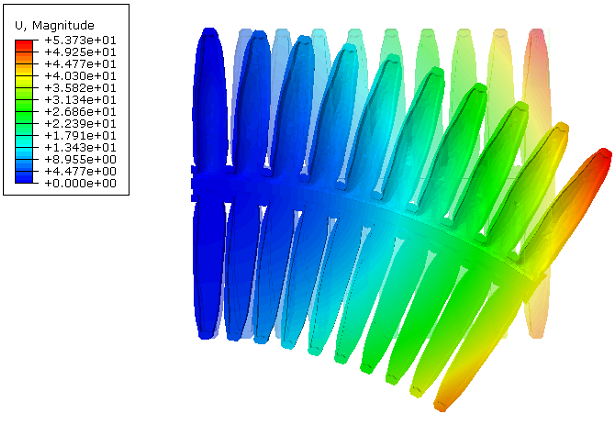}
			\caption*{Top chamber is activated}
		\end{subfigure}
		\hfill
		\begin{subfigure}{0.48\textwidth}
			\centering
			\includegraphics[width=\linewidth]{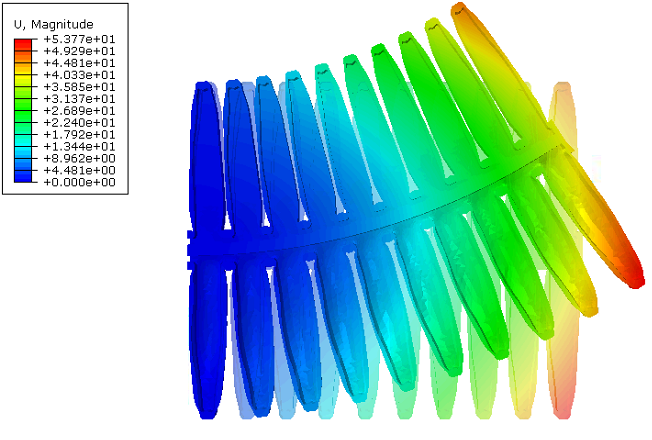}
			\caption*{Bottom chamber is activated}
		\end{subfigure}
		\caption{15 kPa}
	\end{subfigure}
	\hfill

	% ------------------ 20 kPa ------------------
	\begin{subfigure}{0.65\textwidth}
		\centering
		\begin{subfigure}{0.48\textwidth}
			\centering
			\includegraphics[width=\linewidth]{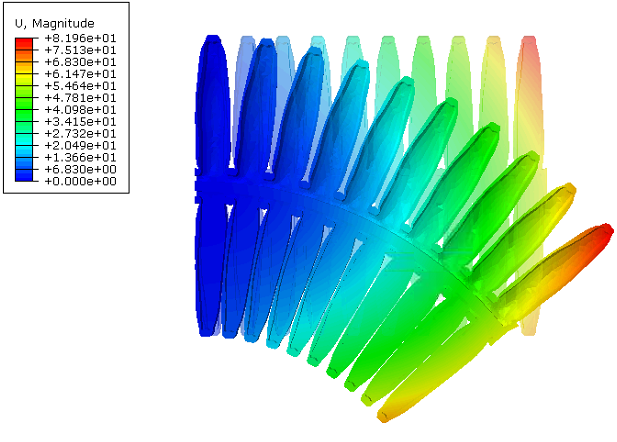}
			\caption*{Top chamber is activated}
		\end{subfigure}
		\hfill
		\begin{subfigure}{0.48\textwidth}
			\centering
			\includegraphics[width=\linewidth]{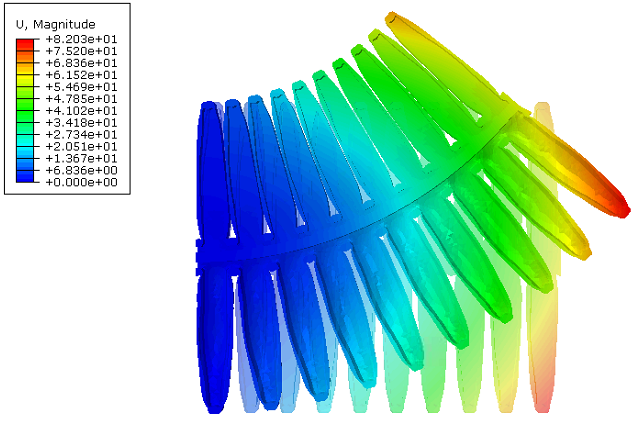}
			\caption*{Bottom chamber is activated}
		\end{subfigure}
		\caption{20 kPa}
	\end{subfigure}

	\caption{Deformation profiles of the optimized soft fish tail showing top and bottom cross-sectional views under different hydraulic pressures (5kPa, 10kPa, 15kPa and 20kPa)}
	\label{fig:Analyses_pressure}
\end{figure*}

\begin{table}[h!]
\centering
\caption{Tip displacement values illustrating the bending response of the soft fish tail under increasing hydraulic pressure from \texttt{ABAQUS} analyses}
\label{tab:Uy_pressure}
\begin{tabular}{c c}
\hline
Pressure (kPa) & $U_y$ (mm) \\
\hline
5  & $-1.12\times10^{-1}$ \\
10 & $-2.87\times10^{0}$ \\
15 & $-5.68\times10^{1}$ \\
20 & $-7.43\times10^{1}$ \\
\hline
\end{tabular}
\end{table}

The optimized soft tail design is depicted in Fig.~\ref{fig:results_2D CAD Model}a. One notices that the optimized chamber differs from the trivial rectangular shape mentioned in~\cite{katzschmann2018exploration}. We perform nonlinear finite element analysis with different pressure loads in \texttt{ABAQUS} to demonstrate how the optimized soft tail outperforms its manually designed counterpart with rectangular chambers. Thus, the optimized tail design can provide better 3D swimming performance than a rectangular-chambered tail. FEA is performed on 2D rectangular and optimized chambers. The 2D-optimized chamber exhibits greater deformation (approximately 17.3\%) under the same hydraulic pressure load (Fig.~\ref{fig:results_2D results}). The 2D CAD model is then extruded to create the corresponding 3D CAD unit, whose top view is as shown in Fig.~\ref{fig:full_3D_CAD_Model}. Lastly, finite element analysis is performed under different hydraulic loads to illustrate the 3D tail's performance~Fig.~\ref{fig:Analyses_pressure} and the corresponding tip deformation in Table~\ref{tab:Uy_pressure}. The Mooney-Rivlin material model is used with the following strain energy function, $W$, as
\begin{equation}
	W = C_{01} \left(\bar{I}_2 -3\right) + C_{10}\left(\bar{I}_1-3\right) + \frac{1}{D_1}(J-1)^2
\end{equation}
where $\bar{I}_1$ and $\bar{I}_2$ are the first and second deviatoric invariants of the right Cauchy--Green tensor, $J$ is the volumetric Jacobian, and $C_{10}$, $C_{01}$, and $D_1$ are material parameters obtained from experimental calibration ($C_{01} = 0.03$, $C_{10}=0.1$ and $D_1 =0.001$ are considered). Self-contact is enabled on all external and internal surfaces of the actuator using a general contact formulation to prevent interpenetration during large deformations. The actuator base is fully clamped by fixing all translational and rotational degrees of freedom, thereby representing a rigid attachment. A symmetry boundary condition is imposed along the longitudinal $z$-symmetry plane to reduce computational cost. The actuator is discretized using ABAQUS \texttt{C3D10HS} quadratic tetrahedral hybrid elements to accurately capture large bending deformations, avoid volumetric locking in the nearly incompressible hyperelastic material, and ensure robust contact behavior.

\section{Conclusions and Future work}\label{Sec:Con}
This paper presents an automated topology optimization method to design a hydraulic soft fish tail for programmable undulating locomotion. The technique uses the Darcy method in conjunction with the drainage term to model the design-dependent behavior of the hydraulic load in the density-based topology optimization setting. A unit of the soft tail, considered as a compliant mechanism activated by hydraulic load, is optimized using a robust formulation that maximizes the output displacement in the desired direction. The volume and strain energy constraints are applied. The latter is employed to obtain a realizable soft-tail design that sustains the applied hydraulic load. The optimized soft-tail design exhibits greater deformation under uniform applied pressure than its trivial rectangular design. An optimized soft tail can help achieve versatile underwater propulsive motion. Nonlinear analysis is performed in \texttt{ABAQUS} using the Mooney-Rivlin material model, and deformed profiles are also demonstrated. Exciting future avenues include 3D fabrication of the optimized fishtail to illustrate the applicability of TO for the synthesis of soft robotic actuators.

\bibliographystyle{ACM-Reference-Format}
\bibliography{sample-base}

\end{document}